\documentclass[10pt]{article} 
\usepackage[preprint]{tmlr}

\usepackage{amsmath,amsfonts,bm}

\def\eqref#1{equation~\ref{#1}}

\def\1{\bm{1}}

\DeclareMathAlphabet{\mathsfit}{\encodingdefault}{\sfdefault}{m}{sl}
\SetMathAlphabet{\mathsfit}{bold}{\encodingdefault}{\sfdefault}{bx}{n}

\usepackage{hyperref}
\usepackage{url}
\usepackage{amssymb}
\usepackage{graphicx}
\usepackage{booktabs}
\makeatletter
\def\@maketitle{\vbox{\hsize\textwidth\centering
{\LARGE\bf\sffamily \@title\par}\vskip \aftertitskip
{\normalsize\bf \@author\par}
\vskip 0.3in minus 0.1in}}
\makeatother
\newcommand{\code}[1]{\texttt{#1}}
\newcommand{\pp}{\,pp}

\title{Measurement Validity and Representation Dynamics of\\Confidence Signals in Large Language Models}

\author{\name Debojeet Ghosh \qquad Shreyansh Kumar \qquad Ayushman Garg}

\begin{document}
\maketitle

\begin{abstract}
Whether a language model ``knows what it knows'' matters for trustworthy deployment and for deciding when it should abstain. Prior work usually measures this through one channel: verbalized confidence, a linear probe on hidden states, or agreement among sampled answers (semantic entropy), and often treats these as interchangeable. We compare all three in a pre-registered audit of five Qwen2.5 models (0.5B--7B, including a base/instruct pair at 7B) on factual retrieval (PopQA, SimpleQA) and multi-step reasoning (GSM8K, MATH). First, probes trained on prefill states to predict whether sampled answers will disagree become informative earlier for retrieval (about 7\% of layer depth on average) than for reasoning (about 28\%). The pre-registered five-layer test supported this only marginally; a finer 21-layer follow-up, designed after those results, supports it more clearly. In most retrieval cells, however, a prompt-only TF-IDF classifier matches the probe, so the early signal may reflect surface features of the question. Second, verbalized confidence is unstable: slightly different prompt templates yield scores that barely correlate. Third, Expected Calibration Error is misleading: verbal confidence looks well calibrated because it clusters near the base rate while discriminating poorly, whereas sampling agreement has worse ECE but about twice the resolution. Of five pre-registered hypotheses, two were falsified, one was partly supported, one was not supported, and one could not be tested because the base model's verbal confidence was degenerate.
\end{abstract}

\section{Introduction}

Large language models (LLMs) demonstrate impressive capabilities across factual retrieval, mathematical problem-solving, and symbolic logic. However, their tendency to produce plausible but factually erroneous statements, often stated with high certainty, poses risks in high-stakes environments such as clinical decision support, legal analysis, and automated code generation. To mitigate these risks, a substantial literature has investigated methods for estimating model confidence to enable selective abstention or routing to human supervisors.

Current approaches to LLM confidence estimation fall predominantly into three methodological paradigms:
\begin{enumerate}
  \item \textbf{Verbalized Elicitation:} Querying the model directly to state its confidence as a probability percentage (0 to 100\%) or qualitative certainty category within the prompt context.
  \item \textbf{Internal Representation Probing:} Training linear classifiers on the model's residual stream hidden representations during the forward pass over the prompt.
  \item \textbf{Behavioral Sampling Consistency:} Generating multiple temperature-perturbed rollouts and measuring the semantic entropy or answer consensus across equivalent clusters.
\end{enumerate}

Empirical studies frequently treat these three channels as interchangeable estimators of a unified epistemic state. In this paper, we conduct a pre-registered, auditable empirical investigation across a 30-cell grid (5 model architectures $\times$ 6 task tiers) to test whether confidence representations are coherent, interchangeable, and mechanistically uniform.

That assumption fails on each dimension we measured. Verbal confidence does not generalize across minor prompt alterations; standard calibration metrics mask large differences in discriminative sorting power; and the depth at which internal probes become informative differs between associative recall and multi-step reasoning tasks, although for recall a prompt-only baseline explains much of the early signal. Raw verbal confidence is nearly uncorrelated with both behavioral and internal signals (pooled $\rho = 0.086$), while behavioral and internal signals agree moderately in most cells (Appendix~\ref{app:corr}).

\section{Related Work}

\paragraph{Verbal Confidence \& Elicitation Fragility.}
Eliciting verbal self-confidence via prompting has been explored extensively \citep{kadavath2022,lin2022,tian2023}. Verbalized confidence is sensitive to elicitation strategy and prompt framing \citep{xiong2024,geng2024}, and auxiliary models can calibrate an LLM from its generations alone \citep{ulmer2024}. Most existing benchmarks evaluate verbal confidence using a single prompt template, implicitly presuming measurement invariance across phrasing.

\paragraph{Probing Hidden Representations.}
Probing internal representations for truthfulness and confidence traces back to the geometry of hidden states \citep{burns2023,azaria2023,marks2023,zou2023}. Hidden states can carry truthfulness and answerability signals that the generated text does not express \citep{li2023,slobodkin2023}. Prior investigations frequently evaluate layer depth on coarse discrete grids (e.g., 25\% increments).

\paragraph{Semantic Entropy \& Behavioral Consistency.}
\citet{kuhn2023} and \citet{farquhar2024} demonstrated that semantic entropy computed across clustered stochastic generations outperforms raw token log-probabilities for hallucination detection, and \citet{wang2023} showed that self-consistency rollouts improve reasoning accuracy. The closest prior work compares verbalized confidence, sequence probability, P(True) and hidden-state probes on shared items \citep{mahaut2024}, and DECK \citep{chauhan2026} proposes a consistency $\times$ confidence taxonomy structurally similar to our quadrant analysis (Section~\ref{sec:h2}). We position this paper as a measurement-validity study rather than a discovery: it reports where a three-channel comparison holds up and where it does not.

\paragraph{Proper Scoring Rules \& Murphy Decomposition.}
In meteorological forecasting and statistical decision theory, \citet{murphy1973}, \citet{brocker2009}, and \citet{gneiting2007} showed that the standard mean squared error of probabilistic forecasts (Brier score) decomposes into three orthogonal components: \emph{Reliability} (calibration error), \emph{Resolution} (sorting power), and \emph{Uncertainty} (inherent task unpredictability). Standard Expected Calibration Error \citep{guo2017,nixon2019} measures only reliability, conflating well-calibrated uninformative forecasts with genuinely discriminative predictions.

\section{Experimental Setup \& Pre-Registered Protocol}

To guard against post-hoc outcome selection, our experimental design, hypotheses H0--H4, pass rules and audit gates were committed in advance in a version-controlled protocol. One analysis in this paper is \emph{not} pre-registered: the 21-tap depth sweep and its 2-consecutive onset rule (Section~\ref{sec:sweep}), which were specified after the pre-registered five-tap results were known. We label it as a follow-up throughout.

\subsection{The 30-Cell Grid Architecture}

The study systematically evaluates five models from the Qwen2.5 open-weights family across six benchmark tiers:
\begin{itemize}
  \item \textbf{Models (5):} Qwen2.5-0.5B-Instruct (24 layers), Qwen2.5-1.5B-Instruct (28 layers), Qwen2.5-3B-Instruct (36 layers), Qwen2.5-7B-Base (28 layers), and Qwen2.5-7B-Instruct (28 layers).
  \item \textbf{Retrieval Tiers (3):} PopQA, top popularity quintile (R1); PopQA, bottom popularity quintile (R2); and SimpleQA (R3; adversarially constructed short-answer questions).
  \item \textbf{Reasoning Tiers (3):} GSM8K (C1; grade-school multi-step arithmetic), MATH levels 1--2 (C2), and MATH levels 4--5 (C3). Each cell holds 500 questions sampled at a fixed seed and split 60/20/20 into train, calibration and test.
\end{itemize}

This gives a grid of 30 cells. Some earlier studies drop cells whose accuracy falls outside a chosen range. We kept all 30 and instead recorded whether each cell's pilot accuracy fell between 25\% and 80\%.

\subsection{Probe Tap Specification \& Architectural Implementation}

We captured activations during a single forward pass over the prompt, before the model generated anything. For each model we read the token embeddings (0\% depth) and the output of each transformer block at the chosen depths.

The probe reads the hidden state at the last prompt token, skipping any padding. The capture was switched on before the prompt was processed and switched off before decoding began, so generated tokens could not leak into the captured representations. An earlier version of the pipeline had this bug: the hooks stayed active and recorded output tokens. It was fixed, and a test in the repository now checks it. Instruct models received prompts formatted with their official chat template, and the base model received raw prompt text.

\subsection{Pre-Registered Safeguards and Gates}

The pipeline enforced four pre-registered audit gates:
\begin{description}
  \item[Gate 1 (Grading Validity).] A human audit of 200 graded outputs, stratified by grader (about 50 each for alias-match, NLI, numeric and symbolic grading), agreed with automated grading on 195/200 (97.5\%; threshold 95\%). All 5 discrepancies were the symbolic grader under-crediting correct math answers because of formatting (4 in C2, 1 in C3, 0 in retrieval), so symbolic-grader agreement was 90\% and reasoning accuracy is slightly understated.
  \item[Gate 2 (Verbal Generalization).] Required pairwise Spearman rank correlation $\rho \ge 0.60$ across three elicitation formats (Format A: integer 0--100; Format B: five ordinal labels; Format C: answer-or-pass decision).
  \item[Gate 3 (Probe Validity).] Probes were trained on the train split and evaluated on the calibration split. A cell passes if some depth reaches AUROC $\ge 0.65$ while beating both a 20-draw label-shuffle null and a prompt-only TF-IDF baseline, and if the $p_0$ negative control holds: the probe at 0\% depth (the token embedding, which carries no question-specific computation at the final prompt token) must score AUROC $0.50 \pm 0.10$. \code{7b-base\_\_C3} failed the negative control ($p_0 = 0.634$) and was dropped.
  \item[Gate 4 (Model Comparison, H3).] Requires matched base/instruct items carrying both a verbal and a behavioral or internal signal. Format parsing was not the problem (the base model parsed on $\ge 99\%$ of items in every format). Its Format B answers, however, used essentially two labels (CERTAIN and NO\_IDEA), so the pre-registered verbal-degeneracy guard ($\ge 3$ distinct values per cell) excluded its verbal signal in all six tiers. With 0 matched rows, H3 declined to test rather than report a misleading 0.0\% rate.
\end{description}

\begin{figure}[t]
  \centering
  \includegraphics[width=0.85\textwidth]{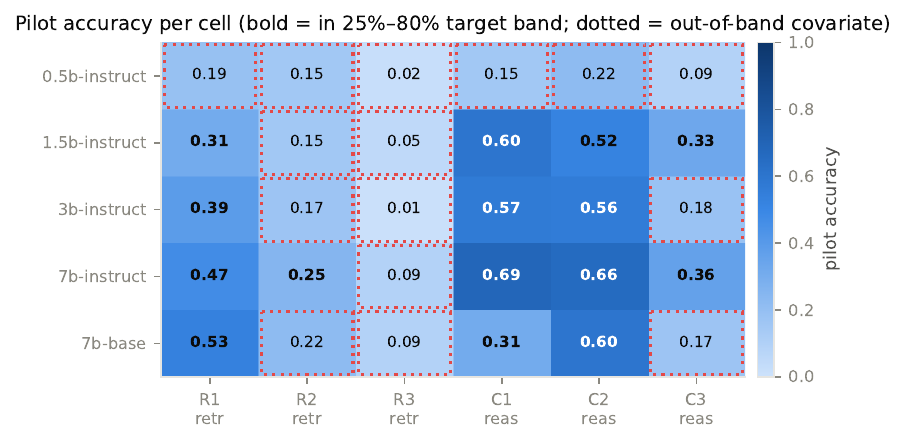}
  \caption{\textbf{Complete 30-cell evaluation matrix.} Pilot accuracy per (model $\times$ tier) cell. Dotted outlines indicate cells outside the 25\% to 80\% accuracy band.}
  \label{fig:grid}
\end{figure}

\section{Representation Depth: Reasoning vs.\ Retrieval (H4)}
\label{sec:h4}

Our mechanistic hypothesis (H4) was that probes begin to predict better than chance at an earlier layer for retrieval than for reasoning, and at an earlier layer in larger models. The pre-registered protocol trains probes on how consistent the model is with itself, not on whether its answer is correct. For each question we sample 10 answers and compute their semantic entropy, which measures how much the answers differ in meaning. A question is labeled ``inconsistent'' if its entropy is above the median for its cell and ``consistent'' otherwise. A probe's onset is the first layer at which it predicts this label better than chance, using the model's state at the last prompt token, before any answer is generated.

\subsection{Continuous 21-Tap Probe Sweeps \& the 2-Consecutive Rule}
\label{sec:sweep}

\paragraph{Pre-registered test.} The protocol specified five depth taps (0, 25, 50, 75, 100\%), a single-point onset (the first tap with AUROC $\ge 0.65$), and a pass rule requiring the bootstrap CI on the reasoning$-$retrieval onset difference to exclude zero. On the coarse run this test passed: $\Delta = +13.64$\pp{} [95\% CI: $+2.27$, $+27.27$] over 22 probe-valid cells. Our internal audit judged it fragile, because onsets could take only three values ($\{25, 50, 75\}$), an exact one-sided permutation test gave $p = 0.052$, and dropping the two C1 cells at 75\% reduced $\Delta$ to 6.06\pp.

\paragraph{Follow-up (not pre-registered).} We therefore re-ran the probe stages over the same generations at 21 depth percentiles (0\%, 5\%, \ldots, 100\%), with a \textbf{2-consecutive confirmation rule} specified and committed before the sweep ran, but after the coarse results were known. A depth percentile $p$ is the onset if:
\begin{equation}
  \text{AUROC}(p) \ge 0.65 \quad \text{and} \quad \text{AUROC}(p + 5\%) \ge 0.65 .
\end{equation}
Across the 28 cells with probe fits (the 0.5B R1 and R2 cells had too few usable rows), 588 logistic-regression probes were fitted (28 cells $\times$ 21 depth percentiles); 24 cells passed Gate 3.

\begin{figure}[t]
  \centering
  \includegraphics[width=\textwidth]{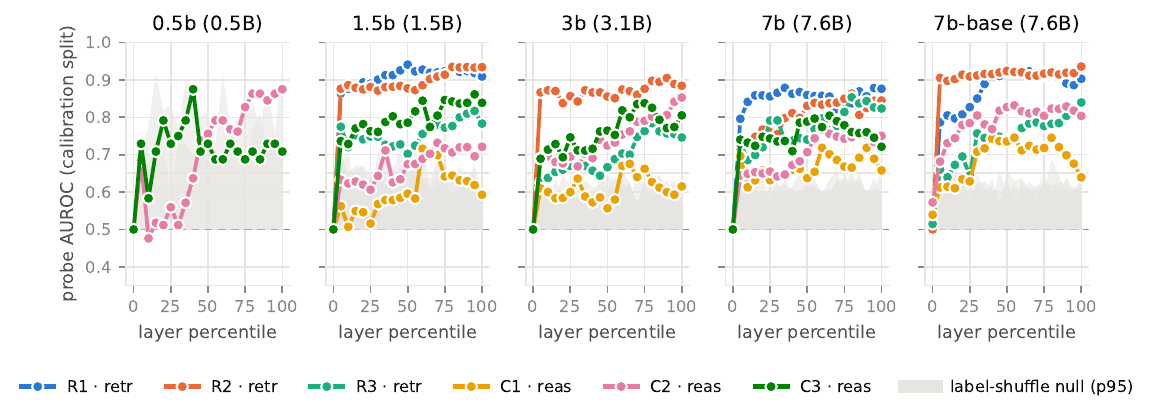}
  \caption{\textbf{AUROC across 21 layer depth percentiles.} Calibration-split AUROC of linear probes predicting the median-split semantic-entropy label from prefill hidden states, faceted by model size. Retrieval tiers (cool lines) mostly cross 0.65 by 5\% depth; reasoning tiers (warm lines) cross later on average. In 9 of 11 retrieval cells the onset-layer probe does not beat the prompt-only TF-IDF baseline (Table~\ref{tab:onsets}).}
  \label{fig:depth}
\end{figure}

\subsection{The Emergence Gap: Full 24-Cell Distribution}

    Table~\ref{tab:onsets} reports \textbf{all 24 cells} that passed Gate 3 in the follow-up sweep (11 retrieval and 13 reasoning cells), together with the prompt-only TF-IDF baseline AUROC and whether the probe at the onset layer beats it.

\begin{table}[t]
  \centering
  \caption{Onset depth for all 24 probe-valid cells (21-tap follow-up sweep).}
  \label{tab:onsets}
  \resizebox{\textwidth}{!}{%
  \begin{tabular}{llrlcccc}
    \toprule
    Model & Tier & Onset (\%) & Onset layer & Probe AUROC (onset) & Probe AUROC (peak) & TF-IDF AUROC & Probe beats TF-IDF at onset \\
    \midrule
    \multicolumn{8}{l}{\textit{Panel A: Factual retrieval tiers ($N = 11$ cells; mean onset $= 6.82\%$, median $= 5.0\%$)}} \\
    Qwen2.5-1.5B-Instruct & R1 (PopQA-high) & 5.0  & Layer 1 / 28 & 0.864 & 0.941 & 0.882 & No \\
    Qwen2.5-1.5B-Instruct & R2 (PopQA-low)  & 5.0  & Layer 1 / 28 & 0.877 & 0.935 & 0.889 & No \\
    Qwen2.5-1.5B-Instruct & R3 (SimpleQA)   & 5.0  & Layer 1 / 28 & 0.775 & 0.817 & 0.728 & Yes \\
    Qwen2.5-3B-Instruct   & R2 (PopQA-low)  & 5.0  & Layer 2 / 36 & 0.867 & 0.905 & 0.868 & No \\
    Qwen2.5-3B-Instruct   & R3 (SimpleQA)   & 15.0 & Layer 5 / 36 & 0.654 & 0.784 & 0.670 & No \\
    Qwen2.5-7B-Base       & R1 (PopQA-high) & 5.0  & Layer 1 / 28 & 0.787 & 0.925 & 0.791 & No \\
    Qwen2.5-7B-Base       & R2 (PopQA-low)  & 5.0  & Layer 1 / 28 & 0.905 & 0.936 & 0.883 & Yes \\
    Qwen2.5-7B-Base       & R3 (SimpleQA)   & 15.0 & Layer 4 / 28 & 0.673 & 0.840 & 0.719 & No \\
    Qwen2.5-7B-Instruct   & R1 (PopQA-high) & 5.0  & Layer 1 / 28 & 0.796 & 0.879 & 0.838 & No \\
    Qwen2.5-7B-Instruct   & R2 (PopQA-low)  & 5.0  & Layer 1 / 28 & 0.697 & 0.863 & 0.788 & No \\
    Qwen2.5-7B-Instruct   & R3 (SimpleQA)   & 5.0  & Layer 1 / 28 & 0.729 & 0.853 & 0.802 & No \\
    \midrule
    \multicolumn{8}{l}{\textit{Panel B: Multi-step reasoning tiers ($N = 13$ cells; mean onset $= 28.08\%$, median $= 30.0\%$)}} \\
    Qwen2.5-1.5B-Instruct & C3 (MATH L4--5) & 5.0  & Layer 1 / 28  & 0.736 & 0.861 & 0.736 & Tie \\
    Qwen2.5-3B-Instruct   & C2 (MATH L1--2) & 5.0  & Layer 2 / 36  & 0.691 & 0.853 & 0.662 & Yes \\
    Qwen2.5-3B-Instruct   & C3 (MATH L4--5) & 5.0  & Layer 2 / 36  & 0.689 & 0.838 & 0.712 & No \\
    Qwen2.5-7B-Base       & C2 (MATH L1--2) & 5.0  & Layer 1 / 28  & 0.682 & 0.832 & 0.695 & No \\
    Qwen2.5-7B-Instruct   & C3 (MATH L4--5) & 5.0  & Layer 1 / 28  & 0.740 & 0.804 & 0.738 & Yes \\
    Qwen2.5-0.5B-Instruct & C3 (MATH L4--5)$^\dagger$ & 15.0 & Layer 4 / 24 & 0.708 & 0.875 & 0.646 & Yes \\
    Qwen2.5-7B-Base       & C1 (GSM8K)      & 30.0 & Layer 8 / 28  & 0.708 & 0.745 & 0.495 & Yes \\
    Qwen2.5-7B-Instruct   & C1 (GSM8K)      & 40.0 & Layer 11 / 28 & 0.659 & 0.718 & 0.627 & Yes \\
    Qwen2.5-7B-Instruct   & C2 (MATH L1--2) & 40.0 & Layer 11 / 28 & 0.672 & 0.767 & 0.716 & No \\
    Qwen2.5-0.5B-Instruct & C2 (MATH L1--2) & 45.0 & Layer 11 / 24 & 0.738 & 0.875 & 0.655 & Yes \\
    Qwen2.5-1.5B-Instruct & C2 (MATH L1--2) & 50.0 & Layer 14 / 28 & 0.675 & 0.732 & 0.592 & Yes \\
    Qwen2.5-1.5B-Instruct & C1 (GSM8K)      & 60.0 & Layer 17 / 28 & 0.716 & 0.716 & 0.519 & Yes \\
    Qwen2.5-3B-Instruct   & C1 (GSM8K)      & 60.0 & Layer 22 / 36 & 0.670 & 0.676 & 0.577 & Yes \\
    \bottomrule
  \end{tabular}}
  \par\smallskip
  \parbox{\textwidth}{\footnotesize $^\dagger$\,\code{0.5b-instruct\_\_C3} was excluded by Gate 3 on the five-tap grid and passes on the 21-tap grid, but its probe rests on 16 calibration and 12 test items and its test-split AUROC at onset is 0.33; treat it as noise. It lowers the reasoning mean, so it works against H4. The other cell that passes only on the fine grid is \code{1.5b-instruct\_\_C1} (onset 60\%). ``Tie'' means the two AUROCs are equal to numerical precision.}
\end{table}

The distributions are:
\begin{itemize}
  \item Retrieval onsets (\code{[5, 5, 5, 5, 5, 5, 5, 5, 5, 15, 15]\%}): nine of eleven cells trigger at 5\% depth (Layer 1 or 2); mean 6.82\%, median 5.0\%.
  \item Reasoning onsets (\code{[5, 5, 5, 5, 5, 15, 30, 40, 40, 45, 50, 60, 60]\%}): mean 28.08\%, median 30.0\%.
  \item Follow-up test (same estimator as the pre-registered rule): $\boldsymbol{\Delta_{\text{mean}} = +21.26}$\textbf{\pp} [95\% bootstrap CI: $+9.72$, $+33.05$]. An exact one-sided permutation test on the difference in means gives $\boldsymbol{p = 0.004}$ (the pipeline's 20,000-draw approximation gives 0.0039). The median gap is $\Delta_{\text{median}} = +25.0$\pp{} (30.0\% vs.\ 5.0\%).
  \item Surface baseline: the onset-layer probe fails to beat the prompt-only TF-IDF baseline in 12 of 24 cells, including 9 of 11 retrieval cells. Early retrieval onsets are therefore not evidence that the network has \emph{computed} anything about recall by 5\% depth; a bag-of-words model of the question predicts the same label about as well.
  \item Scale: the pre-registered prediction that larger models onset earlier is not supported (Spearman $\rho(\text{onset}, \text{parameters}) = -0.22$ [$-0.56$, $+0.21$], $p = 0.30$). Peak AUROC is higher for retrieval than for reasoning (mean 0.88 vs.\ 0.79), as predicted.
\end{itemize}

\subsection{Distributional Nuance}

As shown in Table~\ref{tab:onsets}, five reasoning cells also have onsets at 5\%, all of them MATH cells: \code{1.5b\_\_C3}, \code{3b\_\_C3} and \code{7b-instruct\_\_C3} (levels 4--5), and \code{3b\_\_C2} and \code{7b-base\_\_C2} (levels 1--2). This is not a floor effect: greedy accuracy in these cells ranges from 13\% to 59\%. Two of the five do not beat the TF-IDF baseline at onset. We do not have a confirmed explanation for these early onsets.

By tier, GSM8K (C1) onsets are uniformly late (\code{[30, 40, 60, 60]\%}), MATH levels 1--2 (C2) are mixed (\code{[5, 5, 40, 45, 50]\%}), and MATH levels 4--5 (C3) are early (\code{[5, 5, 5, 15]\%}). The reasoning$-$retrieval gap is therefore carried mainly by C1 and part of C2.

Excluding GSM8K (C1) leaves a smaller gap: $\boldsymbol{\Delta_{\text{no-C1}} = +12.63}$\textbf{\pp} (95\% bootstrap CI [$+1.31$, $+25.76$], 20,000 draws, seed 20260813), which still excludes zero but with a lower bound near it.

\section{Prompt-Format Instability \& Verbal Fragility (H0)}

Hypothesis H0 tested whether verbalized confidence reflects a stable latent property invariant to elicitation formatting ($\rho \ge 0.60$). We elicited verbal confidence across three distinct prompt templates:
\begin{itemize}
  \item \textbf{Format A:} An integer confidence from 0 to 100 that the answer is correct.
  \item \textbf{Format B:} One of five ordinal labels: CERTAIN, FAIRLY\_CONFIDENT, SOMEWHAT\_UNSURE, MOSTLY\_GUESSING, NO\_IDEA. Because the raw Format B value was circular (the empirical accuracy of the chosen label), the canonical verbal signal is $B_{\text{fix}}$, a label-free fixed mapping of these labels to probabilities.
  \item \textbf{Format C:} An answer-or-pass decision under explicit scoring ($+1$ for a correct answer, $-2$ for an incorrect one, 0 for a pass).
\end{itemize}

\begin{table}[h]
  \centering
  \caption{Verbal format correlation matrix (Gate 2 compliance).}
  \label{tab:formats}
  \begin{tabular}{lccc}
    \toprule
    Comparison Pair & Spearman $\rho$ & Pre-Registered Target & Status \\
    \midrule
    Format A vs.\ Format B & $+0.126$ & $\ge 0.60$ & FALSIFIED \\
    Format A vs.\ Format C & $-0.069$ & $\ge 0.60$ & FALSIFIED \\
    Format B vs.\ Format C & $+0.247$ & $\ge 0.60$ & FALSIFIED \\
    \bottomrule
  \end{tabular}
\end{table}

As shown in Table~\ref{tab:formats}, all pairwise correlations were well below the pre-registered 0.60 threshold, and pair A--C was slightly \emph{negative} ($\rho = -0.069$, 95\% CI [$-0.111$, $-0.030$]). Format A is close to a ceiling: 13,121 of 15,000 Format A responses state 90 or higher. Among those, the model still chose to pass in Format C 26.8\% of the time. Verbalized confidence varies with the elicitation format and cannot be treated as a stable scalar property of the model.

\section{Calibration vs.\ Discrimination: Murphy Decomposition (H1)}

Hypothesis H1 pre-registered a calibration ranking, Behavioral $>$ Internal $>$ Verbal, passing only if the best-minus-worst ECE difference had a CI excluding zero.

\subsection{The ECE Equivalence Fallacy}

On the pooled test split of the pre-registered five-tap run, ECE was indistinguishable: Verbal ECE $= 0.0356$ [0.024, 0.061], Behavioral ECE $= 0.0427$ [0.035, 0.061], Internal ECE $= 0.0368$ [0.031, 0.058]. The worst$-$best difference is 0.0071 with a 95\% bootstrap CI that includes zero [$-0.016$, $+0.027$], so H1 is falsified under its pre-registered criterion. (The 21-tap follow-up changes only the internal row: ECE 0.0364, $N = 2{,}400$.) These pooled numbers compare different question sets per signal; Section~\ref{sec:complete} restricts to items carrying all three.

\begin{figure}[t]
  \centering
  \includegraphics[width=\textwidth]{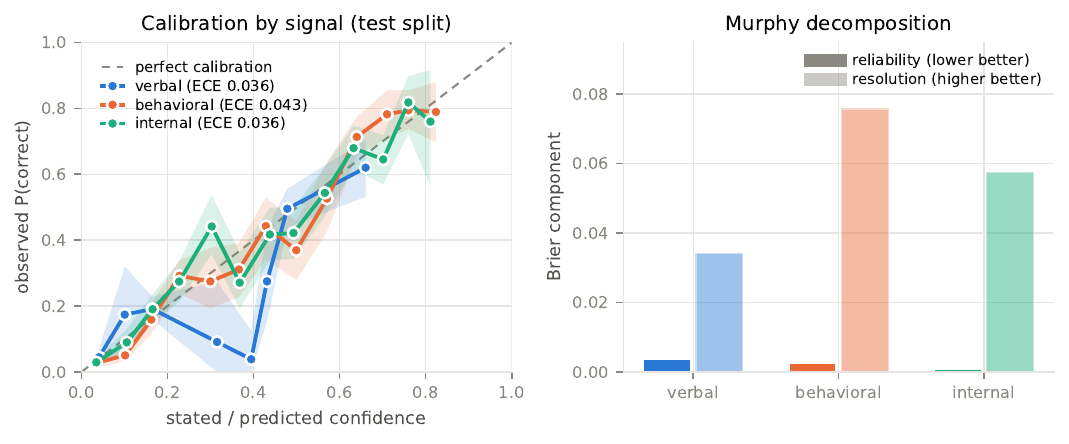}
  \caption{\textbf{Signal calibration and Murphy decomposition on the test split.} \emph{Left:} Calibration curves close to the diagonal; points with $N < 10$ pruned to remove small-sample noise. \emph{Right:} Murphy decomposition of the Brier score. All signals show small reliability error (solid bars), while behavioral consensus achieves more than double the resolution (light bars) of verbal confidence.}
  \label{fig:calibration}
\end{figure}

\subsection{The Complete-Case Tradeoff \& ECE Binning Sensitivity ($N = 842$ Matched Rows)}
\label{sec:complete}

Pooled ECE hides coverage imbalances (follow-up sweep: Verbal $N = 1{,}166$ across 12 cells; Internal $N = 2{,}400$ across 24 cells; Behavioral $N = 2{,}451$ across 29 cells). For a like-for-like comparison, Table~\ref{tab:murphy} uses the $N = 842$ test questions, across 10 cells, where all three signals are present (base rate $\bar{y} = 0.2815$), reporting ECE under 15 uniform bins (pipeline standard) and 10 uniform bins. Reliability and resolution use 10 bins.

\begin{table}[h]
  \centering
  \caption{Complete-case Murphy decomposition and ECE sensitivity ($N = 842$ matched items).}
  \label{tab:murphy}
  \begin{tabular}{lccccc}
    \toprule
    Signal & ECE (15 bins) $\downarrow$ & ECE (10 bins) $\downarrow$ & Reliability $\downarrow$ & Resolution $\uparrow$ & Brier $\downarrow$ \\
    \midrule
    Verbal ($B_{\text{fix}}$) & \textbf{0.0366} & \textbf{0.0326} & \textbf{0.00171} & 0.03160 & 0.1713 \\
    Internal Probe            & 0.0500 & 0.0338 & 0.00219 & 0.05154 & 0.1524 \\
    Behavioral Consensus      & 0.0704 & 0.0680 & 0.00808 & \textbf{0.06371} & \textbf{0.1465} \\
    \bottomrule
  \end{tabular}
\end{table}

Table~\ref{tab:murphy} shows three patterns:
\begin{itemize}
  \item Behavioral consensus scores worse on ECE (0.0704 / 0.0680 vs.\ 0.0366 / 0.0326). A plausible cause is that 10 samples at $T = 0.8$ give a coarse, often polarized agreement score.
  \item Behavioral consensus leads on resolution: 2.02$\times$ that of verbal confidence (0.06371 vs.\ 0.03160) and 1.24$\times$ that of the internal probe (0.05154). This sorting power outweighs the calibration penalty, giving the lowest Brier score (0.1465 vs.\ 0.1524 internal and 0.1713 verbal).
  \item Verbal confidence keeps ECE low by clustering near the base rate: in 9 of the 10 complete-case cells, calibrated verbal confidence never exceeds 0.50. Like a forecaster who always predicts the historical average, it keeps calibration error low while separating correct from incorrect answers poorly.
\end{itemize}

\section{Disagreement Dynamics: Discrete Quadrant Artifact vs.\ Continuous Metric (H2)}
\label{sec:h2}

Hypothesis H2 predicted that disagreements between stated verbal confidence and behavioral/internal confidence fall into distinct quadrants governed by question features, replicating across cells. \textbf{H2 is not supported}: the failing conjunct is replication across cells. We report the quadrant analysis descriptively.

Under the pre-registered 0.5 threshold, the five-tap run gave 90 cases of Excessive Hedging (low verbal, high behavioral) against 9 cases of Performative Certainty (high verbal, low behavioral) among $N = 831$ complete-case items (follow-up sweep: 86 vs.\ 9 among 842). The 10:1 ratio is largely an artifact of the cutoff: in 8 of 9 cells calibrated verbal confidence never crosses 0.5, which suppresses the ``high verbal'' quadrants. The cases are also concentrated: all 9 Performative Certainty items come from one cell (\code{7b-instruct\_\_C2}) and all 90 Excessive Hedging items from two (\code{3b-instruct\_\_C1}, \code{7b-instruct\_\_R1}).

Replacing the cutoff with a continuous mismatch, $\Delta = \text{Verbal} - \text{Behavioral}$, gives a distribution centered at zero: $\text{Mean}(\Delta) = -0.006$, $\text{Median}(\Delta) = -0.0004$, $\text{SD} = 0.114$. Regressing $\Delta$ on question features (length, presence of a year, digits, multiple entities) gives no association (all $p > 0.35$ pooled; all $p > 0.6$ with cell fixed effects). Pooled over cells, quadrant \emph{membership} is associated with some features (e.g., multiple entities, $\chi^2 = 79.6$; digits, $\chi^2 = 68.2$), but no feature reaches significance within any single cell, which is consistent with these features acting as proxies for tier rather than as question-level drivers.

\begin{figure}[t]
  \centering
  \includegraphics[width=\textwidth]{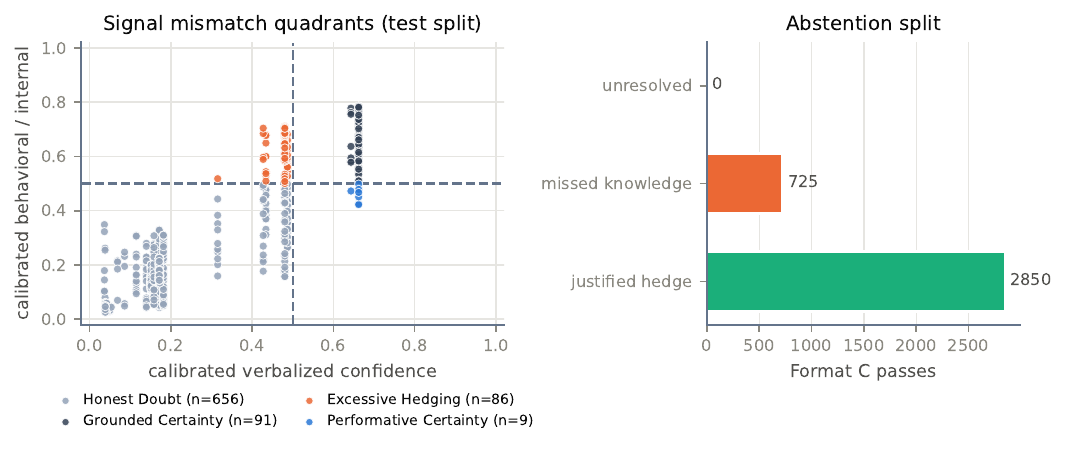}
  \caption{\textbf{Epistemic alignment quadrants and forced-answer abstention behavior.} \emph{Left:} Calibrated verbal vs.\ behavioral/internal confidence. \emph{Right:} The abstention split on Format C passes (justified hedge vs.\ missed knowledge).}
  \label{fig:quadrant}
\end{figure}

\section{Limitations \& Threats to Validity}

\begin{itemize}
  \item \textbf{Single model family.} All evaluations use Qwen2.5 (0.5B to 7B). The depth gap and format fragility need testing on other families.
  \item \textbf{Probe label.} Probes predict a median split of the model's own semantic entropy, not correctness. H4 is about when the prompt representation predicts sampling dispersion, and should not be read as the depth at which the model ``knows it is right''.
  \item \textbf{Surface baseline.} In 9 of 11 retrieval cells, the onset-layer probe does not beat a prompt-only TF-IDF baseline, so early retrieval onsets may reflect surface properties of the question.
  \item \textbf{Pre-registration.} The pre-registered five-tap H4 test passed but was fragile (permutation $p = 0.052$). The 21-tap sweep and 2-consecutive rule are a follow-up designed after those results were seen.
  \item \textbf{Reasoning composition.} The reasoning$-$retrieval gap is carried mainly by GSM8K (C1); MATH levels 4--5 (C3) onsets are early, and excluding C1 roughly halves the gap ($+12.6$\pp, CI lower bound $+1.3$).
  \item \textbf{Grading.} The symbolic grader under-credits correctly formatted math answers (90\% agreement with human labels, all errors one-directional), slightly depressing measured accuracy in C2/C3.
  \item \textbf{Sample size.} 500 questions per cell (100 test items), below the protocol's target of 2,000. Complete-case analyses cover only 9--10 of 30 cells.
  \item \textbf{H3 untested.} Base-vs-instruct alignment could not be evaluated: the 7B base model's verbal confidence was degenerate (essentially two labels) and was excluded in all six tiers.\end{itemize}

\section{Conclusion}

Across 30 experimental cells, the three confidence channels do not behave like interchangeable readings of one quantity. Verbal self-assessment depends on the elicitation format. Sampling consensus has roughly twice the resolution of verbal confidence and the best Brier score, at about 10$\times$ the inference cost. Internal probes become informative at shallower depth for retrieval than for reasoning, but the retrieval signal is largely matched by a prompt-only baseline, and the gap rests on a follow-up analysis that was not pre-registered. On this evidence, practitioners should not route on verbal confidence alone; sampling consensus is the most informative single signal we measured.

\subsubsection*{Broader Impact Statement}
This work evaluates how far confidence signals from LLMs can be trusted for selective abstention. Its main practical implication is cautionary: verbal confidence changes with the elicitation format and should not be used on its own to gate answers in safety-relevant settings, and early-layer probe signals for factual recall are largely matched by a prompt-only baseline. The study uses public benchmarks and open-weight models only, and releases no new model.

\bibliography{references}
\bibliographystyle{tmlr}

\clearpage
\appendix

\section{Correlation Matrix Across Confidence Signals}
\label{app:corr}
\begin{figure}[h]
  \centering
  \includegraphics[width=0.9\textwidth]{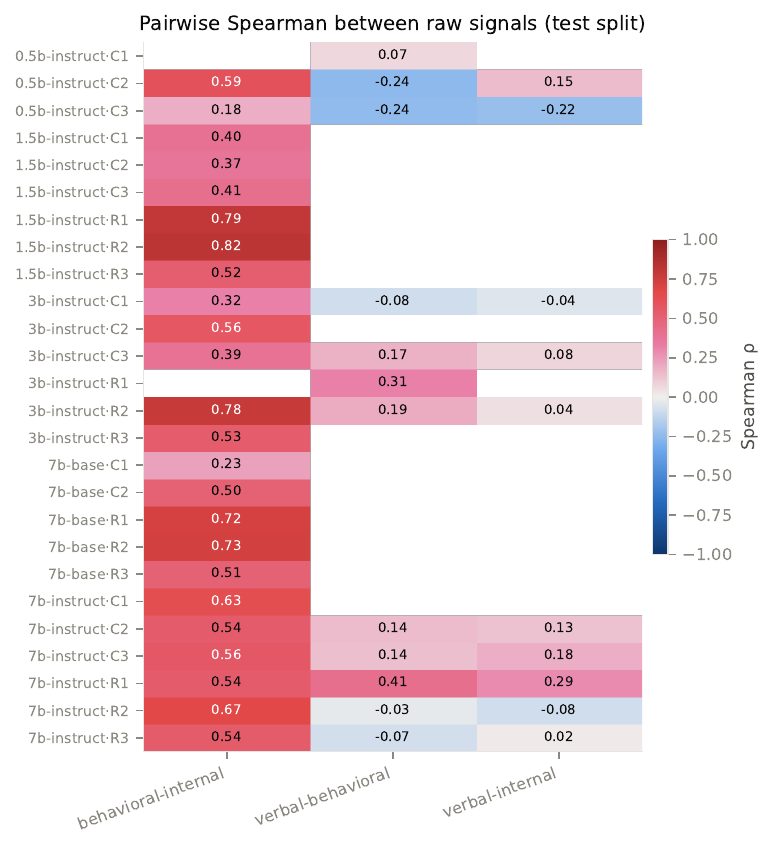}
  \caption{\textbf{Cross-signal rank correlations.} Per-cell Spearman correlations between the raw verbal, behavioral and internal signals, computed where both signals in a pair are present. Pooled over the test split, raw verbal and internal signals correlate at $\rho = 0.086$. After per-cell calibration the pooled correlation rises to 0.83, which mostly reflects differences in difficulty between cells.}
  \label{fig:correlations}
\end{figure}

\section{Compute and Provenance}
\label{app:compute}

All runs used a single NVIDIA RTX PRO 6000 Blackwell GPU. The five-tap generation run did not record the code version it used, but the 21-tap follow up did. Code, raw generations and derived tables are available at \url{https://github.com/PEEXR/Confidence-AI}.

\end{document}